\documentclass[11pt,a4paper]{article}
\usepackage[hyperref]{naaclhlt2018}
\usepackage{times}
\usepackage{latexsym}
\usepackage{graphicx}
\usepackage{amsfonts}
\usepackage{dsfont}
\usepackage{amsmath}

\usepackage{url}

\aclfinalcopy 

\title{Investigations on Knowledge Base Embedding \\ for Relation Prediction and Extraction}

\author{Peng Xu \\
  Department of Computing Science \\
  University of Alberta\\
  Edmonton, Canada\\
  {\tt pxu4@ualberta.ca} \\\And
  Denilson Barbosa \\
  Department of Computing Science \\
  University of Alberta\\
  Edmonton, Canada\\
  {\tt denilson@ualberta.ca} \\}

\date{}

\begin{document}
\maketitle

\begin{abstract}
We report an evaluation of the effectiveness of the existing knowledge base embedding models for relation prediction and for relation extraction on a wide range of benchmarks.
We also describe a new benchmark, which is much larger and complex than previous ones, which we introduce to help validate the effectiveness of both tasks.
The results demonstrate that knowledge base embedding models are generally effective for relation prediction but unable to give improvements for the state-of-art neural relation extraction model with the existing strategies, while pointing limitations of existing methods.
\end{abstract} 

\section{Introduction}
\label{intro}

Representing information about real-world entities and their relations in structured knowledge bases (KBs) enables various applications such as structured search, factual question answering, and intelligent virtual assistants.
A major challenge for using discrete representation of knowledge base is the lack of capability of accessing the similarities among different entities and relations.
Knowledge base embedding (KBE) techniques have been proposed in recent years to deal with this issue.
The main idea is to represent the entities and relations in a vector space, and one can use machine learning technique to learn the continuous representation of the knowledge base in the latent space.

Motivated by the fact that the existing KBs suffer from the problem of low coverage, considerable effort has been committed in automatically deriving new facts to extend a manually built knowledge base with information from the existing knowledge base.
As a result, most of KBE models focus on predicting missing entities or accessing the plausibility of a missing triple instead of predicting missing relations.
Another reason for such choice is that predicting relations is relatively easier than predicting entities because the number of relations is significantly smaller than the number of entities.
Moreover, existing KBE models reach nearly perfect results for relation prediction on some benchmark datasets~\cite{lin2015modeling}.

To the best of our knowledge, the literature lacks a comprehensive effort in validating the \emph{effectiveness} of state-of-the-art KBE models on the task of relation prediction and extraction.
To fill this gap, we report that extensive experiments have been conducted with different KBE models and different datasets for relation prediction.
We choose three KBE models considering their simplicity, effectiveness, and flexibility: {\bf TransE}, {\bf DistMult} and {\bf ComplEx}.
We test on four established benchmarks as well as on a new one we developed, covering different levels of text complexity and corpus size.

\newcite{weston2013connecting} was the first work to combine representation learnt by KBE models and representation learned from textual mentions for relation prediction.
The two representations were trained independently of each other, and were only combined with a simple strategy at inference time.
Since they got remarkable improvement on peformance of relation extraction, we expect that more improvement can also be obtained with the recent development in areas of both KBE and relation extraction.
Following the same strategy, we use the chosen KBE models to help the state-of-art neural models for relation extraction.
In addition, various combining strategies have been tried in order to squeeze the most improvement from KBE models.

\paragraph*{Contribution.}
The experimental results reported here show that the chosen KBE models are enough to achieve satisfactory performances on all of the datasets we studied for relation prediction.
However, the improvements we can squeeze from KBE models are negligible for relation extraction which goes against our expectation.
This observation indicates that the strategies that combine representations at inference time is not effective anymore due to the recent development in areas of both KBE and relation extraction.
New methods should be proposed to make advantage of KBE to facilitate relation extraction.

\section{Related Work}

Recent years have witnessed great advances of knowledge base embedding methods, which represent entities and relations in KBs with vectors or matrices.
{\bf TransE} \cite{bordes2013translating} is the first model to introduce translation-based embedding.
Later variants, such as {\bf TransH} \cite{wang2014knowledge}, {\bf TransR} \cite{lin2015learning} and {\bf TransD} \cite{ji2015knowledge}, extend {\bf TransE} by projecting the embedding vectors of entities into various spaces.
{\bf ManifoldE} \cite{xiao2015one} embeds a triple as a manifold rather than a point.
{\bf RESCAL} \cite{nickel2011three} is one of the earliest studies on embedding based on latent matrix factorization, using a bilinear form as score function.
{\bf DistMult} \cite{yang2014embedding} simplifies {\bf RESCAL} by only using a diagonal matrix, and {\bf Complex} \cite{trouillon2016complex} extends {\bf DistMult} into the complex space.
{\bf HOLE} \cite{nickel2016holographic} employs circular correlation to combine the two entities in a triple.
{\bf ConvE} \cite{dettmers2017convolutional} uses a convolutional neural network as the score function.
However, investigations to validate effectiveness on relation prediction of their models are neglected by most of these studies.

Translation-based models have been widely investigated for their effectiveness to facilitate traditional relation extraction model. 
\cite{weston2013connecting} was the first work to use {\bf TransE} to enrich a text-based model for relation extraction and achieved a significant improvement.
And so do {\bf TransH} \cite{wang2014knowledge}, {\bf TransR} \cite{lin2015learning} and {\bf PTransE} \cite{lin2015modeling}.
However, limited efforts have been committed for latent factor models.
In addition, the models of relation extraction have evolved rapidly in recent years but none of the work tries to utilize knowledge base embeddings to facilitate these state-of-art neural models \cite{zeng2015distant, lin2016neural, wu2017adversarial}.

\section{Models for Knowledge Base Embedding}

For a given knowledge base, let $\mathcal{E}$ be the set of entities with $|\mathcal{E}|=n$, $\mathcal{R}$ be the set of relations with $|\mathcal{R}|=m$, and $\mathcal{T}$ be the set of ground truth triples.
In general, a knowledge base embedding (KBE) model can be formulated as a score function $f_r(s, o), s, o \in \mathcal{E}, r \in \mathcal{R}$ which assigns a score to every possible triple in the knowledge base.
The estimated likelihood of a triple being correct depends only on its score given by the score function.

Different models formulate their score function based on different designs, and therefore interpret scores differently, which further leads to various training objectives.

\subsection{Translation-based Models}

Translation based models are based on the principle first proposed by \cite{bordes2013translating} that if there exists a relationship $r$ between entities $s,o$ then the following relationship between their respective embeddings holds: $\mathbf{e_s} + \mathbf{w_r} \approx \mathbf{e_o}$. The scoring function is thus designed as 

\begin{equation}
f_r(s, o) = \lVert \mathbf{e_s} + \mathbf{w_r} - \mathbf{e_o} \rVert,
\end{equation}

\noindent where $\mathbf{e_s, e_o} \in \mathbb{R}^K$ are entity embedding vectors, $\mathbf{w_r}\in\mathbb{R}^K$ is the relation embedding vector and $K$ is the embedding size. 
A large group of models fall into this category, such as {\bf TransE}, {\bf TransH}, {\bf TransR}, {\bf TransD} and so on, as well as some other models such as {\bf ManifoldE}. 
The objective of training a translation-based model is typically minimizing the following marginal loss:

\begin{equation}
\mathcal{J}_m = \sum_{(s,r,o)\in\mathcal{T}}[\gamma+f_r(s, o)-f_{r'}(s', o')]_+
\end{equation}

\noindent where $[\cdot]_+=\max(0,\cdot)$ is the hinge loss, $\gamma$ is the margin (often set to $1$), and $(s', r', o')$ is a negative triple generated based on the positive triple. 

In this work, we choose {\bf TransE} because of its simplicity and efficiency, as well as the fact other models are extensions of the main idea.
We follow the training procedure of \citet{bordes2013translating} except for generating negative triples. 
Commonly, the negative triple is generated by replacing the subject entity or the object entity of a positive triple with a random entity in the knowledge base. 
In order to capture the relation information better, we also replace the relation of a positive triple with a random relation when generating the negative triples.

\subsection{Latent Factor Models}

Latent factor models assume that the probability of the existence of a tripe $(s, r, o)$ is given by the logistic link function:

\begin{equation}
P((s, r, o)) = \sigma(X_{so}^{(r)})
\end{equation}

\noindent where $X^{(r)} \in \mathbb{R}^{n\times n}$ is a latent matrix of scores of relation $r$. Latent factor models try to find a generic structure for $X^{(r)}$ that leads to a flexible approximation of common relations in real world KBs with matrix factorization.

Based on eigenvalue decomposition $X=EWE^{-1}$, {\bf DistMult} gives a bilinear form as scoring function:

\begin{equation}
f_r(s, o) = \langle \mathbf{e_s}, \mathbf{w_r}, \mathbf{e_o} \rangle.
\end{equation}

\noindent where $\mathbf{w_r} \in \mathbb{R}^K$.
However, this model loses much expressiveness due to its simplicity and cannot describes antisymmetric relations accurately.
In order to handle these issues, {\bf ComplEx} transforms the embeddings of {\bf DistMult} from real space to complex space, which defines the scoring function as:

\begin{equation}
f_r(s, o) = \text{Re}(\langle \mathbf{e_s}, \mathbf{w_r}, \mathbf{e_o} \rangle),
\end{equation}

\noindent where $\mathbf{w_r} \in \mathbb{C}^K$.

There are many other latent factor models, such as {\bf RESCAL}, {\bf HOLE} and many closely related models, like {\bf ConvE}. However, we only choose {\bf DistMult} and {\bf ComplEx} considering their simplicity. In addition, their performance on link prediction are already competitive compared to more sophisticated models.

Models were trained following the same procedure described by \citet{trouillon2016complex} except that we use the same strategy as {\bf TransE} described above to generate the negative triples.

\section{Experiments}

\begin{table*}[ht]
\small
\begin{center}
\begin{tabular}{| l | *{2}{l} | *{2}{l} | *{2}{l} | *{2}{l} | *{2}{l} | *{2}{l} |} 
\hline
& \multicolumn{4}{|c|}{TransE} & \multicolumn{4}{|c|}{DistMult} & \multicolumn{4}{|c|}{ComplEx} \\ \hline
& \multicolumn{2}{|c|}{MRR} & \multicolumn{2}{|c|}{Hits@1} & \multicolumn{2}{|c|}{MRR} & \multicolumn{2}{|c|}{Hits@1} & \multicolumn{2}{|c|}{MRR} & \multicolumn{2}{|c|}{Hits@1} \\ \hline
Dataset & Filter & Raw & Filter & Raw & Filter & Raw & Filter & Raw  & Filter & Raw & Filter & Raw\\ \hline
WN18 & {\bf 0.971} & {\bf 0.969} & {\bf 0.956} & {\bf 0.952} & 0.623 & 0.622 & 0.256 & 0.256 & {\bf 0.991} & {\bf 0.989} & {\bf 0.987} & {\bf 0.983} \\
FB15k & 0.883 & 0.773 & 0.829 & 0.650 & 0.695 & 0.644 & 0.463 & 0.408 & {\bf 0.971} & {\bf 0.840} & {\bf 0.950} & {\bf 0.726} \\ 
WN18RR & 0.843 & 0.842 & 0.735 & 0.734 & 0.871 & 0.866 & 0.810 & 0.802 & {\bf 0.894} & {\bf 0.893} & {\bf 0.813} & {\bf 0.813} \\
FB15k-237 & {\bf 0.955} & {\bf 0.950} & {\bf 0.930} & {\bf 0.921} & 0.926 & 0.921 & 0.880 & 0.871 & {\bf 0.956} & {\bf 0.950} & {\bf 0.933} & {\bf 0.922} \\
FB3M & 0.475 & 0.464 & 0.364 & 0.347 & 0.620 & 0.607 & 0.439 & 0.432 & {\bf 0.683} & {\bf 0.639} & {\bf 0.460} & {\bf 0.410} \\ \hline
\end{tabular}
\caption{Evaluation results on relation prediction.} \label{RP}
\end{center}
\end{table*}

\subsection{Datasets}

To evaluate KBE models for relation prediction, we use four common knowledge base completion datasets from the literature and introduce a new one. 
WN18 \cite{bordes2013translating} is a subset of WordNet which consists of 18 relations and 40,943 entities. 
WN18RR is a subset of WN18 introduced by \cite{dettmers2017convolutional} which removes and dramatically increases the difficulty of reasoning.
FB15k \cite{bordes2013translating} is a subset of Freebase which contains about 15k entities with 1,345 different relations.
Likewise, FB15k-237 is a subset of FB15k introduced by \cite{toutanova2015observed}. FB15k-237 removed redundant relations in FB15k and greatly reduced the number of relations.

To investigate the effectiveness of KBE models to facilitate relation extraction, we use the New York Times corpus (NYT) released by \cite{riedel2010modeling} as training and testing data. 
A new dataset, namely FB3M, is introduced which is also a subset of Freebase restricted to the top 3 million entities - where top is defined as the ones with the largest number of relations to other entities.
This dataset uses a large amount of entities and all possible relationships in Freebase.
Hence, it covers most entities and all the relations to be predicted in the NYT dataset.
Following \cite{weston2013connecting}, we removed all the entity pairs present in the NYT test set from this dataset and translate the deprecated relationships into their new variants.

\subsection{Relation Prediction}

Relation prediction aims to predict relations given two entities. For each testing triple with missing relation, models are asked to compute the scores for all candidate entities and rank them in descending order. 

Following \cite{lin2015modeling}, we use two measures as our evaluation metrics: the mean reciprocal of correct relation ranks (MRR) and the proportion of valid relations ranked in top-1(Hits@1). 
For each metric, we follow evaluation regimes ``Raw" and ``Filter" as described by \citet{bordes2013translating}.

Evaluation results of relation prediction are shown in Table~\ref{RP}. From there we observe that: 
(1) Generally, KBE models are doing well in the task of relation prediction. It indicates that relation information between entities can be captured by the existing KBE models without any specific modifications to adapt this task; 
(2) {\bf ComplEx} achieves the best performance on all datasets. In addition, it significantly and consistently outperforms {\bf DistMult} which matches the observations for the task of link prediction; (2) Surprisingly, {\bf TransE} has competitive performance with {\bf ComplEx} on some datasets which disagrees with the observations for the task of link prediction. It indicates that there exists intrinsic difference between the task of link prediction and relation prediction.

\subsection{Facilitate Relation Extraction}

Relation extraction from text aims to extract relational facts from plain text to enrich existing KBs.
Recent works regard large-scale KBs as source for distant supervision to annotate sentences as training instances and build relation classifiers using neural models \cite{zeng2015distant, lin2016neural, wu2017adversarial}. 
All these methods reason new facts only based on plain text. In this task, we explore the effectiveness of KBE models to facilitate relation extraction from text.

\paragraph*{Combined Model.}
In the experiments, we implemented the RNN-based model presented in \cite{wu2017adversarial}. 
We combine the ranking scores from the neural model with those from KBE to rank testing triples, and generate precision-recall curves for both {\bf TransE} and {\bf ComplEx}. 
Formally, for each pair of entities $(s,o)$ that appear in the test set, all the corresponding sentences $S$ in the test set are collected to form a set $\mathcal{S}$ and a prediction is performed with

\begin{equation}
\hat r_{s,o}= \underset{r \in \mathcal{R}}{\operatorname{argmax}}\ S_{RE}(r|\mathcal{S})
\end{equation}

\noindent where $S_{RE}(r|\mathcal{S})$ indicates the plausibility of relation $r$ given the sentences set $\mathcal{S}$ predicted by the neural model. The predicted relation can either be a valid relation or NA - a marker that means there is no relation between $s$ and $o$ (NA is added to $\mathcal{R}$ during training and is treated like other relations). If $\hat r_{s,o}$ is a relation, a composite score is defined:

\begin{align}
S(s, \hat r_{s, o}, o)=&\alpha S_{RE}(\hat r_{s, o}|\mathcal{S}) \nonumber\\& +\nonumber\\& (1-\alpha) f_{\hat r_{s,o}}(s, o)
\end{align}

\noindent where $\alpha\in(0, 1]$ is a hyper-parameter to tune the balance between the text information and the KB. 
That is, only the top scoring non-NA predictions are re-scored. Hence, our final composite model favors predictions that agree with both the text information and the KB. 
If $\hat r_{s,o}$ is NA, the score is unchanged. If $\alpha=1$, it's the same as the neural relation extraction model; if $\alpha=0.5$, it follows the same combining strategy described in \cite{weston2013connecting}.

\begin{figure}[h]
\begin{center}
 \includegraphics[height=2in]{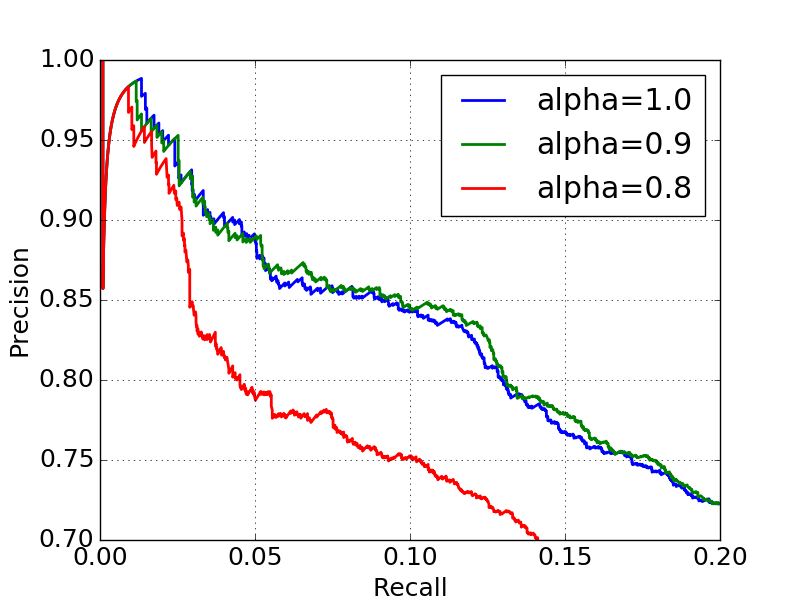}
\end{center}
\caption{Precision-recall curves of {\bf TransE} with different alpha.}
\label{transe}
\end{figure}

\begin{figure}[h]
\begin{center}
 \includegraphics[height=2in]{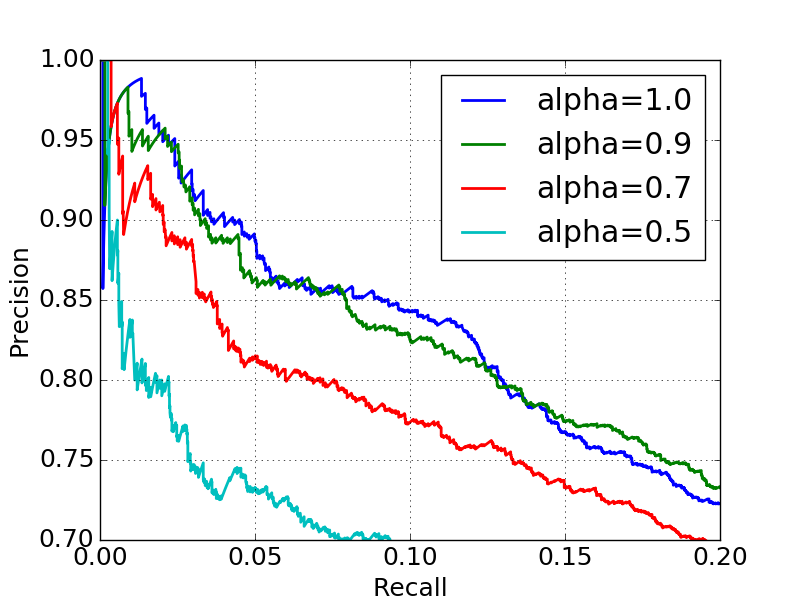}
\end{center}
\caption{Precision-recall curves of {\bf ComplEx} with different alpha.}
\label{complex}
\end{figure}

The evaluation curves for {\bf TransE} and {\bf ComplEx} are shown in Figure~\ref{transe} and~\ref{complex}.
We can see that the improvement is negligible when $\alpha=0.9$ and the combined models perform consistently worse than the original neural model with smaller $\alpha$.

We also tried other strategies for combining the models:
(1) use geometric average, harmonic average instead of weighted average; 
(2) apply the average operation on all non-NA predictions or all predictions instead of just the top non-NA predictions; 
(3) transform the scores into a probability distribution by applying a \emph{softmax} operation. However, all these changes make no improvement or sometimes hurt the performance.

The experimental results indicate that the existing strategies which perform the combining at inference time is not effective any more due to the recent development in areas of both KBE and relation extraction.
New methods should be proposed to make better advantage of KBE to facilitate relation extraction.

\section{Conclusion and Further Work}

In this paper, we investigate the effectiveness of the existing knowledge base embedding models on two tasks: relation predication and relation extraction.
Experimental results show that the existing combining strategies cannot yield expected improvements for relation extraction, while the existing KBE models can achieve almost perfect performance on relation prediction.
It points one direction of our study, that is, seeking new approaches to effectively leverage the structured information encoded by the KBE models for relation extraction. 



%



\bibliography{krl4rc}
\bibliographystyle{acl_natbib}

\end{document}